\documentclass{article}

\PassOptionsToPackage{authoryear, round}{natbib}

\usepackage[final]{neurips_2021}

\usepackage[utf8]{inputenc} 
\usepackage[T1]{fontenc}    
\usepackage{url}            
\usepackage{booktabs}       
\usepackage{amsfonts}       
\usepackage{nicefrac}       
\usepackage{microtype}      
\usepackage[dvipsnames]{xcolor}         
\usepackage[colorlinks, citecolor=Maroon]{hyperref}       
\usepackage{wrapfig}
\usepackage{array}
\usepackage{amsmath}
\usepackage{amssymb}
\usepackage{graphicx}
\usepackage{multirow}
\usepackage{tabularx}
\usepackage{subcaption}
\usepackage{xspace}

\newcommand{\R}{\mathbb{R}}
\newcommand{\f}[1]{f_\text{#1}}
\newcommand{\loss}[1]{L_\text{#1}}
\newcommand{\norm}[1]{\lVert{#1}\rVert}

\newcommand{\ie}{\textit{i.e.}\xspace}
\newcommand{\eg}{\textit{e.g.}\xspace}

\newcommand{\eref}[1]{(\ref{#1})}
\newcommand{\secref}[1]{Sec.~\ref{#1}}
\newcommand{\figref}[1]{Fig.~\ref{#1}}
\newcommand{\tabref}[1]{Tab.~\ref{#1}}

\newcommand\blfootnote[1]{%
  \begingroup
  \renewcommand\thefootnote{}\footnote{#1}%
  \addtocounter{footnote}{-1}%
  \endgroup
}

\title{NeRS: Neural Reflectance Surfaces \\ for Sparse-view 3D Reconstruction in the Wild}

\author{
    Jason Y. Zhang \hspace{5mm} Gengshan Yang \hspace{5mm} Shubham Tulsiani$^*$ \hspace{5mm} Deva Ramanan$^*$\\
    Robotics Institute, Carnegie Mellon University
 
}

\begin{document}

\maketitle

\begin{abstract}
Recent history has seen a tremendous growth of work exploring implicit representations of geometry and radiance, popularized through Neural Radiance Fields (NeRF).  Such works are fundamentally based on a (implicit) {\em volumetric} representation of occupancy, allowing them to model diverse scene structure including translucent objects and atmospheric obscurants. But because the vast majority of real-world scenes are composed of well-defined surfaces, we introduce a {\em surface} analog of such implicit models called Neural Reflectance Surfaces (NeRS). NeRS learns a neural shape representation of a closed surface that is diffeomorphic to a sphere, guaranteeing water-tight reconstructions. Even more importantly, surface parameterizations allow NeRS to learn (neural) bidirectional surface reflectance functions (BRDFs) that factorize view-dependent appearance into environmental illumination, diffuse color (albedo), and specular “shininess.” Finally, rather than illustrating our results on synthetic scenes or controlled in-the-lab capture, we assemble a novel dataset of multi-view images from online marketplaces for selling goods. Such “in-the-wild” multi-view image sets pose a number of challenges, including a small number of views with unknown/rough camera estimates. We demonstrate that surface-based neural reconstructions enable learning from such data, outperforming volumetric neural rendering-based reconstructions. We hope that NeRS serves as a first step toward building scalable, high-quality libraries of real-world shape, materials, and illumination. The project page with code and video visualizations can be found at \href{https://jasonyzhang.com/ners}{jasonyzhang.com/ners}. \blfootnote{$^*$~~denotes equal coding. Corresponding email: \href{mailto:jasonyzhang@cmu.edu}{jasonyzhang@cmu.edu}.}
\end{abstract}

\begin{figure*}[ht]
    \centering
    \includegraphics[width=\textwidth]{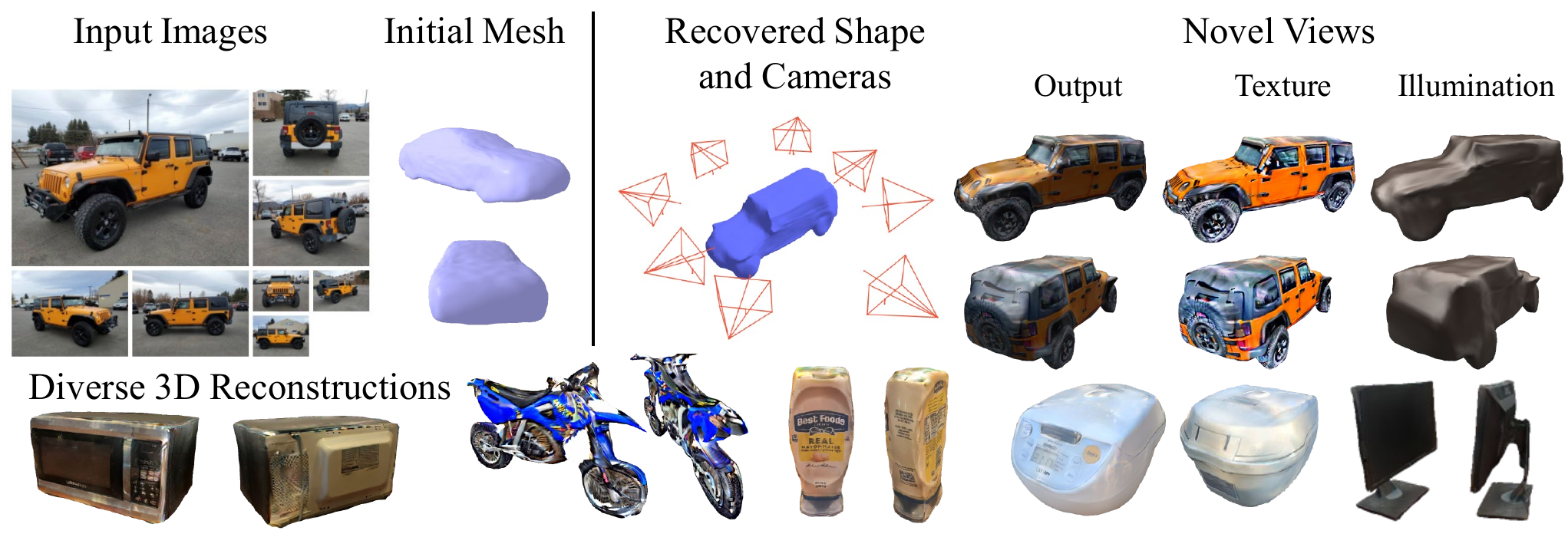}
    \caption{\textbf{3D view synthesis in the wild.} From several multi-view internet images of a truck and a coarse initial mesh (top left), we recover the camera poses, 3D shape, texture, and illumination (top right). We demonstrate the scalability of our approach on a wide variety of indoor and outdoor object categories (second row). \href{https://jasonyzhang.com/ners/paper_figures\#fig1}{[Video]}}
    \label{fig:teaser}
\end{figure*}

\section{Introduction}

Although we observe the surrounding world only via 2D percepts, it is undeniably 3D. The goal of recovering this underlying 3D from 2D observations has been a longstanding one in the vision community, and any computational approach aimed at this task must answer a central question about representation---how should we model the geometry and appearance of the underlying 3D structure? 

An increasingly popular answer to this question is to leverage neural \emph{volumetric} representations of density and radiance fields~\citep{mildenhall2020nerf}. This allows modeling structures from rigid objects to translucent fluids, while further enabling arbitrary view-dependent lighting effects. However, it is precisely this unconstrained expressivity that makes it less robust and unsuitable for modeling 3D objects from sparse views in the wild. While these neural volumetric representations have been incredibly successful, they require hundreds of images, typically with precise camera poses, to model the full 3D structure and appearance of real-world objects. In contrast, when applied to `in-the-wild' settings \eg a sparse set of images with imprecise camera estimates from off-the-shelf systems (see \figref{fig:teaser}), they are unable to infer a coherent 3D representation. We argue this is because these neural volumetric representations, by allowing arbitrary densities and lighting, are \emph{too} flexible.

Is there a robust alternative that captures real-world 3D structure? The vast majority of real-world objects and scenes comprise of well-defined \emph{surfaces}. This implies that the geometry, rather than being an unconstrained volumetric function, can be modeled as a 2D manifold embedded in euclidean 3D space---and thus encoded via a (neural) mapping from a 2D manifold to 3D. Indeed, such meshed surface manifolds form the heart of virtually all rendering engines~\citep{foley1996computer}.  Moreover, instead of allowing arbitrary view-dependent radiance, the appearance of such surfaces can be described using (neural) bidirectional surface reflection functions (BRDFs), themselves developed by the computer graphics community over decades. We operationalize these insights into \emph{Neural Reflectance Surfaces} (NeRS), a surface-based neural representation for geometry and appearance.

NeRS represents shape using a neural displacement field over a canonical sphere, thus constraining the geometry to be a watertight surface. This representation crucially associates a surface normal to each point, which enables modeling view-dependent lighting effects in a physically grounded manner. Unlike volumetric representations which allow unconstrained radiance, NeRS factorizes surface appearance using a combination of diffuse color (albedo) and specularity. It does so by learning neural texture fields over the sphere to capture the albedo at each surface point, while additionally inferring an environment map and surface material properties. This combination of a surface constraint and a factored appearance allows NeRS to learn efficiently and robustly from a sparse set of images in the wild, while being able to capture varying geometry and complex view-dependent appearance.

Using only a coarse category-level template and approximate camera poses, NeRS can reconstruct instances from a diverse set of classes. Instead of evaluating in a synthetic setup, we introduce a dataset sourced from marketplace settings where multiple images of a varied set of real-world objects under challenging illumination are easily available. We show NeRS significantly outperforms neural volumetric or classic mesh-based approaches in this challenging setup, and as illustrated in \figref{fig:teaser}, is able to accurately model the view-dependent appearance via its disentangled representation. 
Finally, as cameras recovered in the wild are only approximate, we propose a new evaluation protocol for \textit{in-the-wild} novel view synthesis in which cameras can be refined during both training \textit{and} evaluation.
We hope that our approach and results highlight the several advantages that neural surface representations offer, and that our work serves as a stepping stone for future investigations.

\section{Related Work}

{\bf Surface-based 3D Representations.} As they enable efficient representation and rendering, polygonal meshes are widely used in vision and graphics. In particular, morphable models~\citep{blanz1999morphable} allow parametrizing shapes as deformations of a canonical template and can even be learned from category-level image collections~\citep{cashman2012shape,kar2015category}. With the advances in differentiable rendering~\citep{kato2018neural,Laine2020diffrast,ravi2020pytorch3d}, these have also been leveraged in learning based frameworks for shape prediction~\citep{cmrKanazawa18,gkioxari2019,ucmrGoel20} and view synthesis~\citep{riegler2020free}. Whereas these approaches use an explicit discrete mesh, some recent methods have proposed using continuous neural surface parametrization like ours to represent shape~\citep{groueix2018papier} and texture~\citep{tulsiani2020implicit,bhattad2021view}.

However, all of these works leverage such surface representations for (coarse) single-view 3D prediction given a category-level training dataset. In contrast, our aim is to infer such a representation given multiple images of a single instance, and without prior training. Closer to this goal of representing a single instance in detail, contemporary approaches have shown the benefits of using videos~\citep{vmr2020,yang2021lasr} to recover detailed shapes, but our work tackles a more challenging setup where correspondence/flow across images is not easily available. In addition, while these prior approaches infer the surface texture, they do not enable the view-dependent appearance effects that our representation can model.

{\bf Volumetric 3D and Radiance Fields.} Volumetric representations for 3D serve as a common, and arguably more flexible alternative to surface based representations, and have been very popular for classical multi-view reconstruction approaches~\citep{furukawa2015multi}. These have since been incorporated in deep-learning frameworks for shape prediction~\citep{girdhar2016learning,choy20163d} and differentiable rendering~\citep{yan2016perspective,tulsiani2017multi}. Although these initial approaches used discrete volumetric grids, their continuous neural function analogues have since been proposed to allow finer shape~\citep{mescheder2019occupancy,park2019deepsdf} and texture modeling~\citep{oechsle2019texture}.

Whereas the above methods typically aimed for category-level shape representation, subsequent approaches have shown particularly impressive results when using these representations to model a single instance from images~\citep{sitzmann2019deepvoxels,thies2019deferred,sitzmann2019srns} -- which is the goal of our work.  More recently, by leveraging an implicit representation in the form of a Neural Radiance Field, \cite{mildenhall2020nerf} showed the ability to model complex geometries and illumination from images. There has since been a flurry of impressive work to further push the boundaries of these representations and allow modeling deformation~\citep{pumarola2020d,park2020deformable}, lighting variation~\citep{martin2020nerf}, and similar to ours, leveraging insights from surface rendering to model radiance~\citep{yariv2020multiview,boss2021nerd,oechsle2021unisurf,srinivasan2021nerv,nerfactor,wang2021neus,yariv2021volume}. However, unlike our approach which can efficiently learn from a sparse set of images with coarse cameras, these approaches rely on a dense set of multi-view images with precise camera localization to recover a coherent 3D structure of the scene. While recent concurrent work~\citep{lin2021barf} does relax the constraint of precise cameras, it does so by foregoing view-dependent appearance, while still requiring a dense set of images. Other approaches~\citep{bi2020deep,zhang2021neural} that learn material properties from sparse views require specialized illumination rigs.

{\bf Multi-view Datasets.} Many datasets study the longstanding problem of multi-view reconstruction and view synthesis. However, they are often captured in controlled setups, small in scale, and not diverse enough to capture the span of real world objects. Middlebury~\citep{seitz2006comparison} benchmarks multi-view reconstruction, containing two objects with nearly Lambertian surfaces. DTU~\citep{aanaes2016large} contains eighty objects with various materials but is still captured in a lab with controlled lighting. Freiburg cars~\citep{SB15} captures 360 degree videos of fifty-two outdoor cars for multi-view reconstruction. ETH3D~\citep{Schops_2019_CVPR} and Tanks and Temples~\citep{knapitsch2017tanks} contain both indoor and outdoor scenes but are small in scale. Perhaps most relevant are large-scale datasets of real-world objects such as Redwood~\citep{Choi2016} and Stanford Products~\citep{oh2016deep}, but the data is dominated by single-views or small baseline videos. In contrast, our Multi-view Marketplace Cars (MVMC) dataset contains thousands of multi-view captures of in-the-wild objects under various illumination conditions, making it suitable for studying and benchmarking algorithms for multi-view reconstruction, view synthesis, and inverse rendering.

\section{Method}
\label{sec:method}

Given a sparse set of input images of an object under natural lighting conditions, we aim to model its shape and appearance. While recent neural volumetric approaches share a similar goal, they require a dense set of views with precise camera information. Instead, our approach relies only on approximate camera pose estimates and a coarse category-level shape template. Our key insight is that instead of allowing unconstrained densities popularly used for volumetric representations, we can enforce a \emph{surface}-based 3D representation. Importantly, this allows view-dependent appearance variation by leveraging constrained reflection models that decompose appearance into diffuse and specular components. In this section, we first introduce our (neural) surface representation that captures the object's shape and texture, and then explain how illumination and specular effects can be modeled for rendering. Finally, we describe how our approach can learn using challenging in-the-wild images.

\begin{figure}[t]
    \centering
    \includegraphics[width=\textwidth]{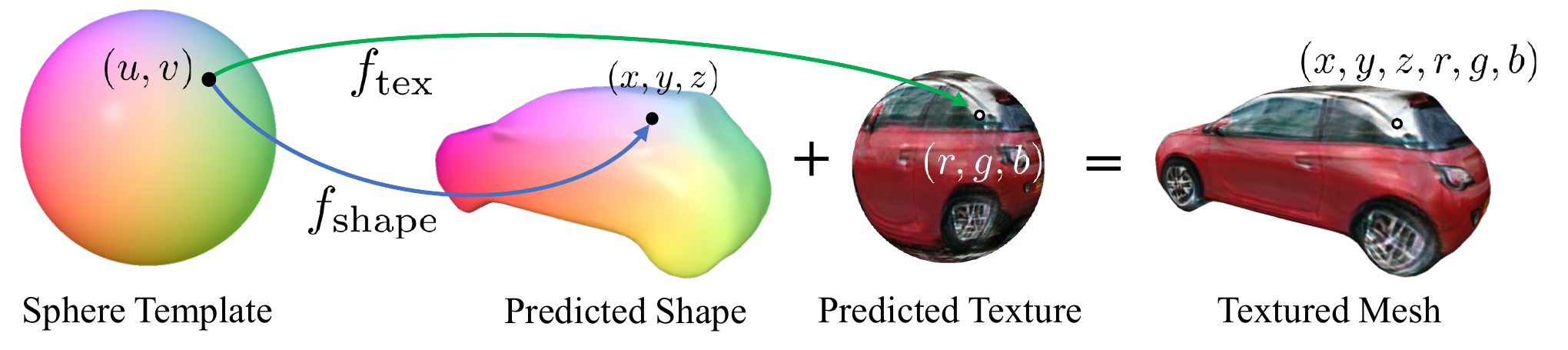}
    \caption{\textbf{Neural Surface Representation.} We propose an implicit, continuous representation of shape and texture. We model shape as a deformation of a unit sphere via a neural network $\f{shape}$, and texture as a learned per-uv color value via a neural network $\f{tex}$. We can discretize $\f{shape}$ and $\f{tex}$ to produce the textured mesh above.}
    \label{fig:surface_representation}
\end{figure}

\subsection{Neural Surface Representation}
\label{sec:surface}

We represent object shape via a deformation of a unit sphere. Previous works \citep{cmrKanazawa18,ucmrGoel20} have generally modeled such deformations \textit{explicitly}: the unit sphere is discretized at some resolution as a 3D mesh with $V$ vertices. Predicting the shape deformation thus amounts to predicting vertex offsets $\delta \in \R^{V\times 3}$. Such \textit{explicit discrete} representations have several drawbacks. First, they can be computationally expensive for dense meshes with fine details. Second, they lack useful spatial inductive biases as the vertex locations are predicted independently. Finally, the learned deformation model is fixed to a specific level of discretization, making it non-trivial, for instance, to allow for more resolution as needed in regions with richer detail. These limitations also extend to texture  parametrization commonly used for such discrete mesh representations---using either per-vertex or per-face texture samples \citep{kato2018neural}, or fixed resolution texture map, limits the ability to capture finer details.

\begin{wrapfigure}{R}{0.3\textwidth}
\centering
\vspace{-20pt}
    \includegraphics[width=0.29\textwidth]{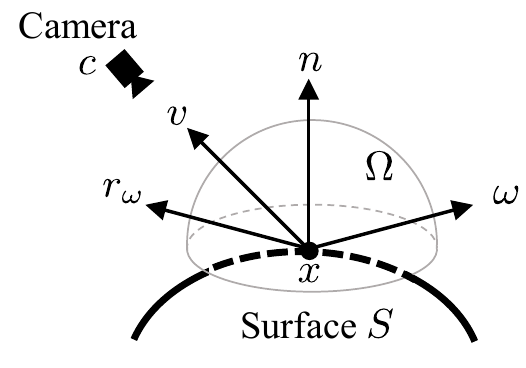}
    \caption{\textbf{Notation and convention for viewpoint and illumination parameterization.} The camera at $c$ is looking at point $x$ on the surface $S$. $v$ denotes the direction of the camera w.r.t $x$, and $n$ is the normal of $S$ at $x$. $\Omega$ denotes the unit hemisphere centered about $n$. We compute the light arriving in the direction of every $\omega \in \Omega$, and $r$ is the reflection of $w$ about $n$.}
    \label{fig:light_model}
    \vspace{-25pt}
\end{wrapfigure}

Inspired by \cite{groueix2018papier,tulsiani2020implicit}, we address these challenges by adopting a continuous surface representation via a neural network. We illustrate this representation in \figref{fig:surface_representation}. For any point $u$ on the surface of a unit sphere $\mathbb{S}^2$, we represent its 3D deformation $x\in \R^3$ using the mapping $\f{shape}(u) = x$ where $\f{shape}$ is parameterized as a multi-layer perceptron.
This network therefore induces a deformation field over the surface of the unit sphere, and this deformed surface serves as our shape representation. We represent the surface texture in a similar manner--as a neural vector field over the surface of the sphere: $\f{tex}(u) = t \in \R^3$. This surface texture can be interpreted as an implicit UV texture map.

\subsection{Modeling Illumination and Specular Rendering}
\label{sec:illumination}

\paragraph{Surface Rendering.}
The surface geometry and texture are not sufficient to infer appearance of the object \eg a uniformly red car may appear darker on one side, and lighter on the other depending on the direction of incident light. In addition, depending on viewing direction and material properties, one may observe different appearance for the same 3D point \eg shiny highlight from certain viewpoints. More formally, assuming that a surface does not emit light, the outgoing  radiance $L_o$ in direction $v$ from a surface point $x$ can be described by the rendering equation \citep{kajiya1986rendering, immel1986radiosity}:
\begin{equation}
    L_o(x, v) = \int_\Omega f_r(x, v, \omega) L_i(x, \omega) (\omega \cdot n)d\omega
    \label{eq:rendering_equation}
\end{equation}
where $\Omega$ is the unit hemisphere centered at surface normal $n$, and $\omega$ denotes the negative direction of incoming light. $f_r(x, v, \omega)$ is the bidirectional reflectance function (BRDF) which captures material properties (\eg color and shininess) of surface $S$ at $x$, and $L_i(x, \omega)$ is the radiance coming toward $x$ from $\omega$ (Refer to \figref{fig:light_model}). Intuitively, this integral computes the total effect of the reflection of every possible light ray $\omega$ hitting $x$ bouncing in the direction $v$.

We thus need to infer the environment lighting and surface material properties to allow realistic renderings. However, learning arbitrary lighting $L_i$ or reflection models $f_r$ is infeasible given sparse views, and we need to further constrain these to allow learning. Inspired by concurrent work \citep{wu2021derender} that demonstrated its efficacy when rendering rotationally symmetric objects, we leverage the Phong reflection model \citep{phong1975illumination} with the lighting represented as a neural environment map.

\paragraph{Neural Environment Map.} An environment map intuitively corresponds to the assumption that all the light sources are infinitely far away. This allows a simplified model of illumination, where the incoming radiance only depends on the direction $\omega$ and is independent of the position $x$ \ie $L_i(x,\omega) \equiv I_\omega$. We implement this as a neural spherical environment map $\f{env}$ which learns to predict the incoming radiance for any query direction 
    $L_i(x,\omega) \equiv I_\omega = \f{env}(\omega)$.
Note that there is a fundamental ambiguity between material properties and illumination, \eg a car that appears red could be a white car under red illumination, or a red car under white illumination. To avoid this, we follow \cite{wu2021derender}, and further constrain the environment illumination to be grayscale, \ie $\f{env}(\omega) \in \R$.

\begin{figure}[t]
    \centering
    \includegraphics[width=\textwidth]{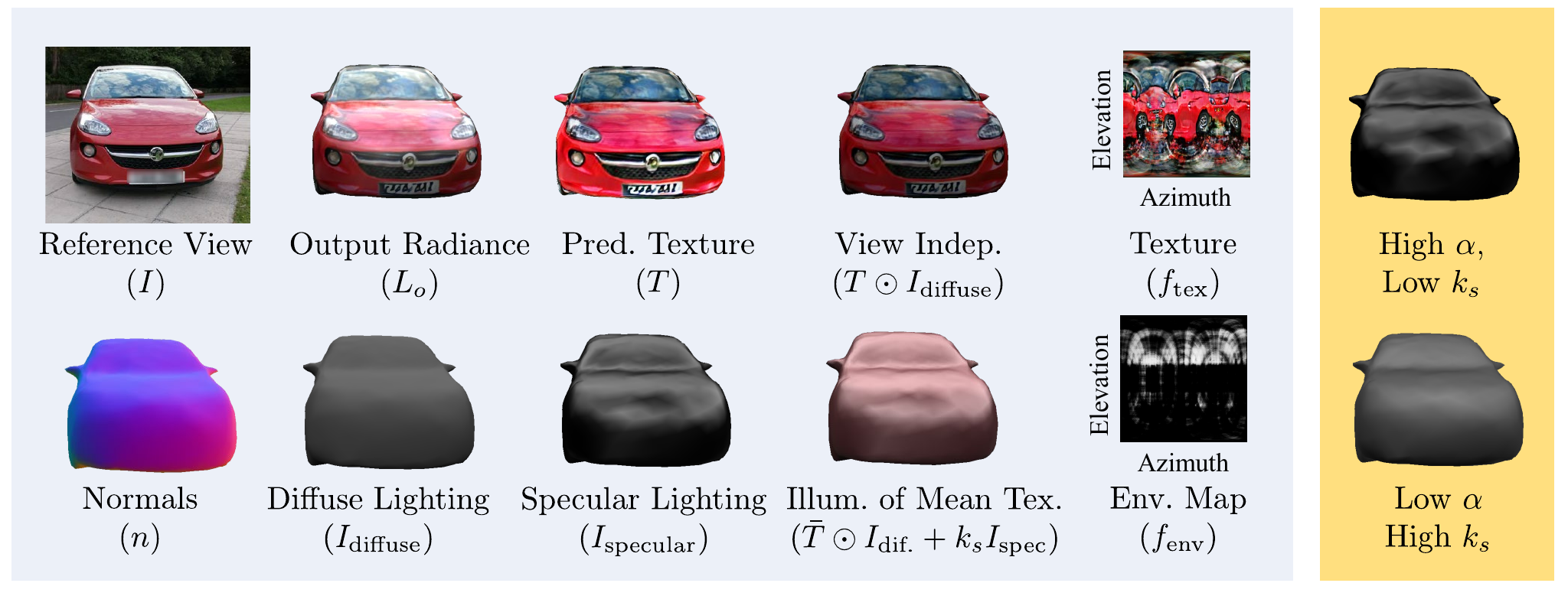}
    \caption{\textbf{Components of learned illumination model.} Given a query camera viewpoint (illustrated via the reference image $I$), we recover the radiance output $L_o$, computed using Phong shading \citep{phong1975illumination}. Here, we show the full decomposition of learned components. From the environment map $\f{env}$ and normals $n$, we compute  diffuse ($I_\text{diffuse}$) and specular lighting ($I_\text{specular}$). The texture and diffuse lighting form the view-independent component (``View Indep.") and the specular lighting (weighted by the specular coefficient $k_s$) forms the view-dependent component of the radiance. Altogether, the output radiance $L_o=T \odot I_\text{diffuse} + k_s I_\text{specularity}$ \eref{eq:radiance}. We also visualize the radiance using the mean texture, which is used to help learn plausible illumination.
    In the yellow box, we visualize the effects of the two specularity parameters. The shininess $\alpha$ controls the mirror-ness/roughness of the surface. The specular coefficient $k_\text{s}$ controls the intensity of the specular highlights.}
    \label{fig:illumination}
\end{figure}

\paragraph{Appearance under Phong Reflection.}
Instead of allowing an arbitrary BRDF $f_r$, the Phong reflection model decomposes the outgoing radiance from point $x$ in direction $v$ into the diffuse and specular components.
The \textit{view-independent} portion of the illumination is modeled by the diffuse component:
\begin{equation}
    I_\text{diffuse}(x) = \sum_{\omega\in \Omega} (\omega \cdot n)I_\omega,
\end{equation}
while the \textit{view-dependent} portion of the illumination is modeled by the specular component:
\begin{equation}
        I_\text{specular}(x, v) = \sum_{\omega\in \Omega} (r_{\omega,n} \cdot v)^\alpha I_\omega,
\end{equation}
where $r_{\omega,n} = 2(\omega \cdot n)n - \omega$ is the reflection of $\omega$ about the normal $n$. The shininess coefficient $\alpha\in (0, \infty)$ is a property of the surface material and controls the ``mirror-ness" of the surface. If $\alpha$ is high, the specular highlight will only be visible  if $v$ aligns closely with $r_\omega$. Altogether, we compute the radiance of $x$ in direction $v$ as:
\begin{equation}
        \label{eq:radiance}
        L_o(x, v) = T(x) \cdot I_\text{diffuse}(x) + k_s \cdot I_\text{specular}(x, v)
\end{equation}
where the specularity coefficient $k_\text{s}$ is another surface material property that controls the intensity of the specular highlight. $T(x)$ is the texture value at $x$ computed by $\f{tex}$. For the sake of simplicity, $\alpha$ and $k_s$ are shared across the entire instance. See \figref{fig:illumination} for a full decomposition of these components.

\subsection{Learning NeRS in the Wild}

\begin{figure}[t]
    \centering
    \includegraphics[width=\textwidth]{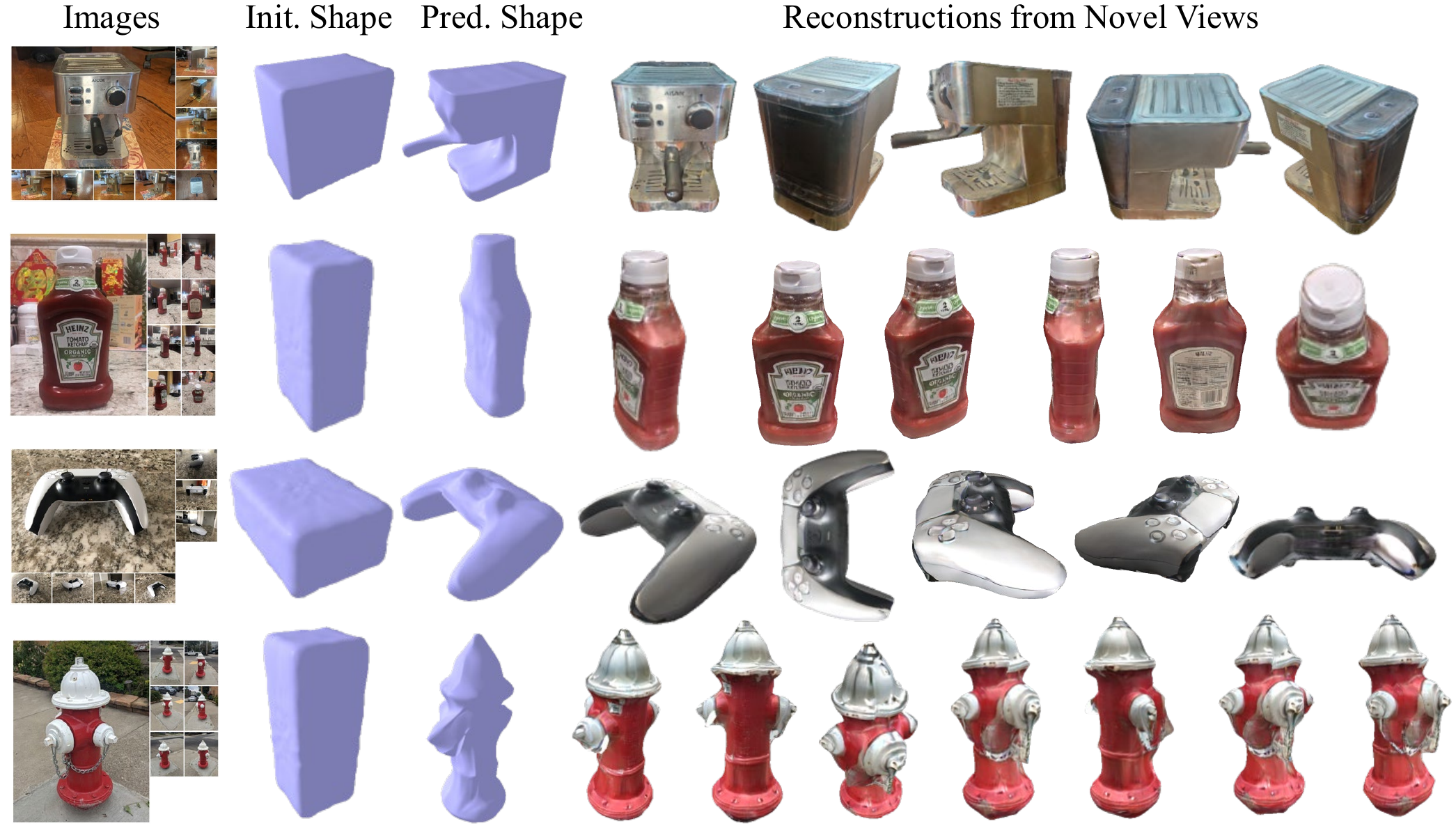}
    \caption{\textbf{Qualitative results on various household objects.} We demonstrate the versatility of our approach on an espresso machine, a bottle of ketchup, a game controller, and a fire hydrant. Each instance has 7-10 input views. We find that a coarse, cuboid mesh is sufficient as an initialization to learn detailed shape and texture. We initialize the camera poses by hand, roughly binning in increments of 45 degrees azimuth. \href{https://jasonyzhang.com/ners/paper_figures\#fig5}{[Video]}}
    \label{fig:qualitative_misc}
\end{figure}

Given a sparse set of images in-the-wild, our approach aims to infer a NeRS representation, which when rendered, matches the available input.
Concretely, our method takes as input $N$  (typically 8) images of the same instance $\{I_i\}_{i=1}^N$, noisy camera rotations $\{R_i\}_{i=1}^N$, and a category-specific mesh initialization $\mathcal{M}$. Using these, we aim to optimize full perspective cameras $\{\Pi\}_{i=1}^N$ as well as the neural surface shape $\f{shape}$, surface texture $\f{text}$, and environment map $\f{env}$. In addition, we also recover the material properties of the object, parametrized by a specularity coefficient $k_s$ and shininess coefficient $\alpha$.

{\bf Initialization.} Note that both the camera poses and mesh initialization are only required to be coarsely accurate. We use an off-the-shelf approach \citep{Xiao2019PoseFromShape} to predict camera rotations, and we find that a cuboid is sufficient as an initialization for several instances (See \figref{fig:qualitative_misc}). We use off-the-shelf approaches~\citep{rother2004grabcut,kirillov2020pointrend} to compute masks $\{M_i\}_{i=1}^N$.
We assume that all images were taken with the same camera intrinsics. We initialize the shared global focal length $f$ to correspond to a field of view of 60 degrees, and set the principal point at center of each image.
We initialize the camera pose with the noisy initial rotations $R_i$ and a translation $t_i$ such that the object is fully in view. We pre-train $\f{shape}$ to output the template mesh $\mathcal{M}$.

{\bf Rendering.} To render an image, NeRS first discretizes the neural shape model $\f{shape}(u)$ over spherical coordinates $u$ to construct an explicit triangulated surface mesh. This triangulated mesh and camera $\Pi_i$ are fed into PyTorch3D's differentiable renderer \citep{ravi2020pytorch3d} to obtain per-pixel (continuous) spherical coordinates and associated surface properties:
\begin{equation}
[UV,N,{\hat M}_i] = \text{Rasterize}(\pi_i,\f{shape})
\label{eq:rasterize}
\end{equation}
where $UV[p]$, $N[p]$, and $\hat{M}[p]$ are (spherical) uv-coordinates, normals, and binary foreground-background labels corresponding to each image pixel $p$.
Together with the environment map $\f{env}$ and specular material parameters $(\alpha,k_s)$, these quantities are sufficient to compute the outgoing radiance at each pixel $p$ under camera viewpoint $\Pi_i$ using \eqref{eq:radiance}. In particular, denoting by $v(\Pi, p)$ the viewing direction for pixel $p$ under camera $\Pi$, and using $u \equiv UV[p], n \equiv N[p]$ for notational brevity, the intensity at pixel $p$ can be computed as:
\begin{equation}
\hat{I}[p] = \f{tex}(u) \cdot \Big(\sum_{\omega\in \Omega} (\omega \cdot n)~ \f{env}(\omega)\Big) + k_s \Big(\sum_{\omega\in \Omega} (r_{\omega,n}\cdot v(\Pi,p))^a  ~\f{env}(\omega)\Big)
\label{eq:radiance_calculation}
\end{equation}
{\bf Image loss.}  We compute a perceptual loss \citep{zhang2018perceptual} $\loss{perceptual}(I_i, \hat{I}_i)$ that compares the distance between the rendered and true image using off-the-shelf VGG deep features. Note that being able to compute a perceptual loss is a significant benefit of surface-based representations over volumetric approaches such as NeRF~\citep{mildenhall2020nerf}, which operate on batches of rays rather than images, due to the computational cost of volumetric rendering. 
Similar to \cite{wu2021derender}, we find that an additional rendering loss using the mean texture (see \figref{fig:illumination} and \figref{fig:qualitative} for examples with details in the supplement) 
helps learn visually plausible lighting.

{\bf Mask Loss.} To measure disagreement between the rendered and measured silhouettes, we compute a mask loss $ \loss{mask} = \frac{1}{N}\sum_{i=1}^N \norm{M_i - \hat{M}_i}_2^2$, distance transform loss $\loss{dt} = \frac{1}{N}\sum_{i=1}^N D_i \odot \hat{M}_i$, and 2D chamfer loss $\loss{chamfer} = \frac{1}{N}\sum_{i=1}^{N}\sum_{p \in E(M_i)}\min_{\hat{p} \in \hat{M_i}} \norm{p - \hat{p}}_2^2$. $D_i$ refers to the Euclidean distance transform of mask $M_i$, $E(\cdot)$ computes the 2D pixel coordinates of the edge of a mask, and $\hat{p}$ is every pixel coordinate in the predicted silhouette.

{\bf Regularization.} Finally, to encourage smoother shape whenever possible, we incorporate a mesh regularization loss $\loss{regularize} = \loss{normals} + \loss{laplacian}$ consisting of normals consistency and Laplacian smoothing losses \citep{nealen2006laplacian,desbrun1999implicit}. Note that such geometry regularization is another benefit of surface representations over volumetric ones. Altogether, we minimize:
\begin{equation}
    L = \lambda_1\loss{mask} + \lambda_2\loss{dt} + \lambda_3\loss{chamfer} + \lambda_4\loss{perceptual} + \lambda_5\loss{regularize}
    \label{eq:total_loss}
\end{equation}
w.r.t $\Pi_i=\begin{bmatrix} R_i, t_i, f\end{bmatrix}$, $\alpha$, $k_s$, and the weights of $\f{shape}$, $\f{text}$, and $\f{env\_map}$.

{\bf Optimization.} We optimize \eref{eq:total_loss} in a coarse-to-fine fashion, starting with a few parameters and slowly increasing the number of free parameters. We initially optimize \eref{eq:total_loss}, w.r.t only the camera parameters $\Pi_i$. After convergence, we sequentially optimize $\f{shape}$, $\f{tex}$, and $\f{env}/\alpha/k_s$. 
We find it helpful to sample a new set of spherical coordinates $u$ each iteration when rasterizing. This helps propagate gradients over a larger surface and prevent aliasing.
With 4 Nvidia 1080TI GPUs, training NeRS requires approximately 30 minutes. Please see \secref{sec:appendix} for hyperparameters and additional details.

\section{Evaluation}
\label{sec:eval}

In this section, we demonstrate the versatility of Neural Reflectance Surfaces to recover meaningful shape, texture, and illumination from in-the-wild indoor and outdoor images.

\textbf{Multi-view Marketplace Dataset.} To address the shortage of in-the-wild multi-view datasets, we introduce a new dataset, Multi-view Marketplace Cars (MVMC), collected from an online marketplace with thousands of car listings. Each user-submitted listing contains seller images of the same car instance.
In total, we curate a subset of size 600 with at least 8 exterior views (averaging 10 exterior images per listing) along with 20 instances for an evaluation set (averaging 9.1 images per listing).
We use \cite{Xiao2019PoseFromShape} to compute rough camera poses.
MVMC contains a large variety of cars under various illumination conditions (\eg indoors, overcast, sunny, snowy, etc). 
The filtered dataset with anonymized personally identifiable information (\eg license plates and phone numbers), masks, initial camera poses, and optimized NeRS cameras will be made available on the \href{https://jasonyzhang.com/ners}{project page}.

\textbf{Novel View Synthesis.}
Traditionally, novel view synthesis requires accurate target cameras to use as queries. Existing approaches use COLMAP~\citep{schoenberger2016sfm} to recover ground truth cameras, but this consistently fails on MVMC due to specularities and limited views.
On the other hand, we can use learning-based methods~\citep{Xiao2019PoseFromShape} to recover camera poses for both training and test views. 
However, as these are inherently approximate, this complicates training \textit{and} evaluation. To account for this, we explore two evaluation protocols. First, to mimic the traditional evaluation setup, we obtain pseudo-ground truth cameras (with manual correction) and freeze them during training and evaluation. While this evaluates the quality of the 3D reconstruction, it does not evaluate the method's ability to jointly recover cameras. As a more realistic setup for evaluating view synthesis in the wild, we evaluate each method with approximate (off-the-shelf) cameras, while allowing them to be optimized.

\textbf{Novel View Synthesis \textit{with Fixed Cameras}.}  
In the absence of ground truth cameras, we create pseudo-ground truth by manually correcting cameras recovered by jointly optimizing over all images for each object instance. For each evaluation, we treat one image-camera pair as the target and the remaining pairs for training. We repeat this process for each image in the evaluation set (totaling 182). 
Unless otherwise noted, qualitative results use approximate cameras and not the pseudo-ground truth.

\begin{figure}[t]
    \centering
    \includegraphics[width=\textwidth]{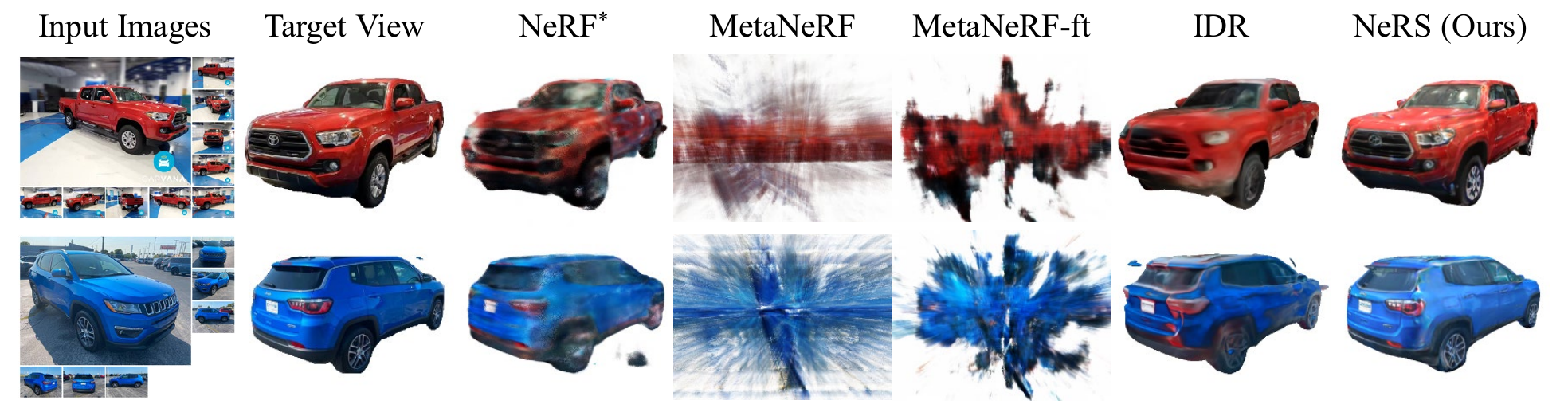}
    \caption{\textbf{Qualitative comparison \textit{with fixed cameras}.} We evaluate all baselines on the task of novel view synthesis on Multi-view Marketplace Cars trained and tested with fixed, pseudo-ground truth cameras. One image is held out during training. Since we do not have ground truth cameras, we treat the optimized cameras from optimizing over all images as the ground truth cameras. We train a modified version (See \secref{sec:eval}) of NeRF~\citep{mildenhall2020nerf} that is more competitive with sparse views (NeRF$^*$). We also evaluate against a meta-learned initialization of NeRF with and without finetuning until convergence~\citep{tancik2020meta}, but found poor results perhaps due to the domain shift from Shapenet cars. Finally, IDR~\citep{yariv2020multiview} extracts a surface from an SDF representation, but struggles to produce a view-consistent output given limited input views.  We find that NeRS synthesizes novel views that are qualitatively closer to the target. The red truck has 16 total views while the blue SUV has 8 total views. \href{https://jasonyzhang.com/ners/paper_figures\#fig6}{[Video]}}
    \label{fig:comparison}
\end{figure}

\begin{figure}[t]
    \centering
    \includegraphics[width=\textwidth]{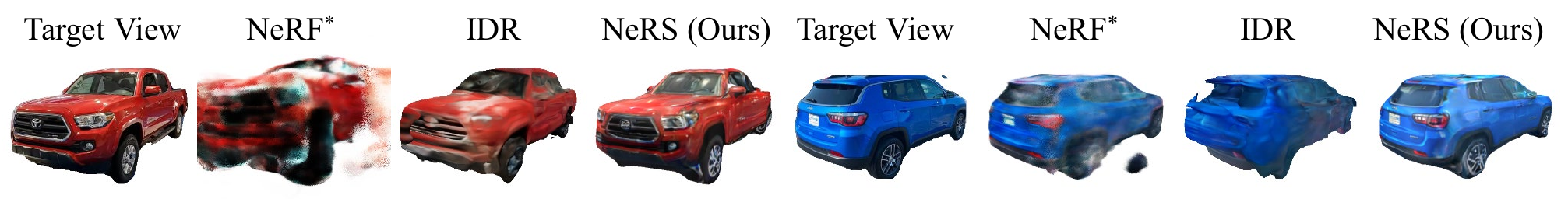}
    \caption{\textbf{Qualitative results for \textit{in-the-wild} novel view synthesis.} Since off-the-shelf camera poses are only approximate for both training and test images, we allow cameras to be optimized during both training and evaluation (See \tabref{tab:full_results_trained} and \secref{sec:eval}). We find that NeRS generalizes better than the baselines in this unconstrained but more realistic setup.  \href{https://jasonyzhang.com/ners/paper_figures\#fig7}{[Video]}}
    \label{fig:comparison_trained}
\end{figure}

\begin{figure}[t]
    \centering
    \includegraphics[width=\textwidth]{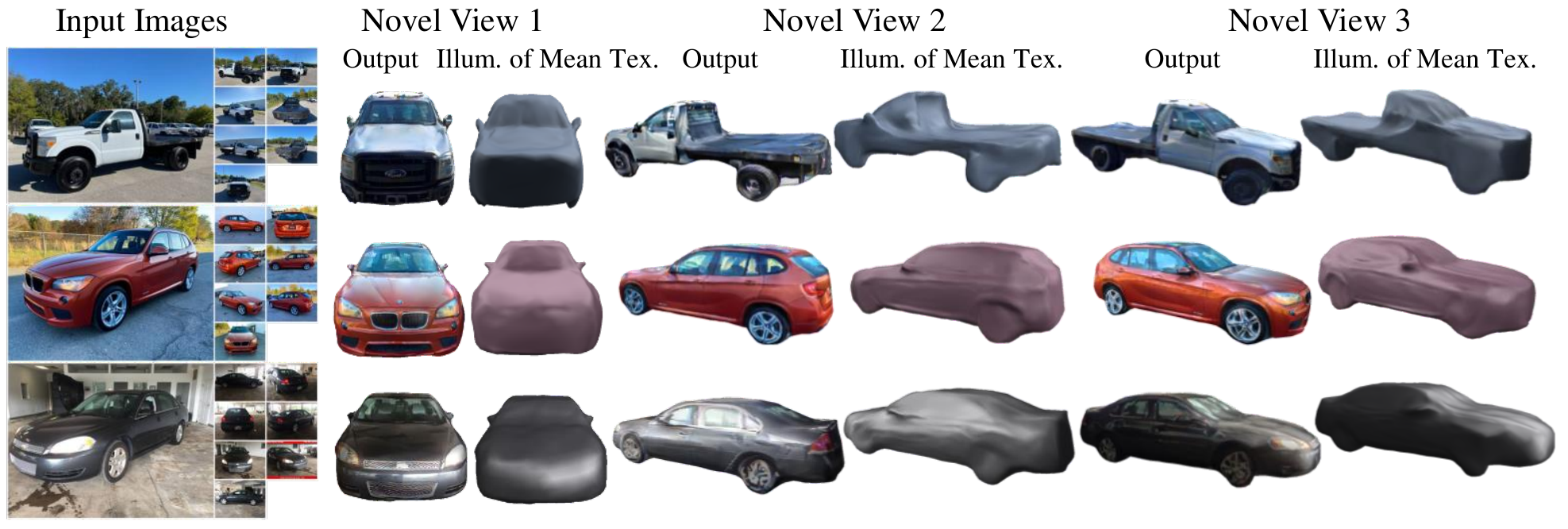}
    \caption{\textbf{Qualitative results on our in-the-wild Multi-view Marketplace Cars dataset.} Here we visualize the NeRS outputs as well as the illumination of the mean texture on 3 of listings from the MVMC dataset. We find that NeRS recovers detailed textures and plausible illumination. Each instance has 8 input views. \href{https://jasonyzhang.com/ners/paper_figures\#fig8}{[Video]}}
    \label{fig:qualitative}
\end{figure} 
\begin{table}[t]
    \centering
    \small{
    \begin{tabular}{l c c c c c}
        \toprule
        Method & MSE $\downarrow$ & PSNR $\uparrow$ & SSIM $\uparrow$ & LPIPS $\downarrow$ & FID $\downarrow$ \\
        \cmidrule{1-6}
        NeRF$^*$~\citep{mildenhall2020nerf} & 0.0393 & 16.0 &  0.698 &  0.287 & 231.7\\
        MetaNeRF~\citep{tancik2020meta}& 0.0755 & 11.4 &  0.345 &  0.666 & 394.5\\
        MetaNeRF-ft~\citep{tancik2020meta} & 0.0791 & 11.3  & 0.500 & 0.542 & 326.8 \\
        IDR~\citep{yariv2020multiview} & 0.0698  & 13.8  & 0.658 &  0.328 & 190.1\\
        NeRS (Ours)  & \textbf{0.0254} & \textbf{16.5} &  \textbf{0.720} & \textbf{0.172} & \textbf{60.9}\\
        \bottomrule\\
    \end{tabular}
    }
    \caption{\textbf{Quantitative evaluation of novel-view synthesis on MVMC using \textit{fixed pseudo-ground truth cameras}.} To evaluate novel view synthesis in a manner consistent with previous works that assume known cameras, we obtain pseudo-ground truth cameras by manually correcting off-the-shelf recovered cameras. 
    We evaluate against a modified NeRF (NeRF$^*$), a meta-learned initialization to NeRF with and without finetuning (MetaNeRF), and the volumetric surface-based IDR. NeRS significantly outperforms the baselines on all metrics on the task of novel-view synthesis with fixed cameras. See \figref{fig:comparison} for qualitative results.}
    \vspace{-7pt}
    \label{tab:full_results_fixed}
\end{table}

\begin{table}[t!]
    \centering
    \small{
    \begin{tabular}{l c c c c c}
        \toprule
        Method & MSE $\downarrow$ & PSNR $\uparrow$ & SSIM $\uparrow$ & LPIPS $\downarrow$ & FID $\downarrow$ \\
        \cmidrule{1-6}
        NeRF$^*$~\citep{mildenhall2020nerf} & 0.0464 & 14.7 &  0.660 &  0.335 & 277.9\\
        IDR~\citep{yariv2020multiview} & 0.0454 & 14.4  & 0.685 &  0.297 & 242.3\\
        NeRS (Ours)  & \textbf{0.0338} & \textbf{15.4} &  \textbf{0.675} & \textbf{0.221} & \textbf{92.5}\\
        \bottomrule\\
    \end{tabular}
    }
    \caption{\textbf{Quantitative evaluation of \textit{in-the-wild} novel-view synthesis on MVMC.}
    Off-the-shelf cameras estimated for in-the-wild data are inherently erroneous. This means that both training and test cameras are approximate, complicating training \textit{and} evaluation.
    To compensate for approximate test cameras, we allow methods to refine the test camera given
    the test image with the model fixed (see \secref{sec:appendix} for details). Intuitively this measures the ability of a method to synthesize a test image under \textit{some} camera.
    We evaluate against NeRF and IDR, and find that NeRS outperforms the baselines across all metrics. See \figref{fig:comparison_trained} for qualitative results.}
    \vspace{-7pt}
    \label{tab:full_results_trained}
\end{table}

\textbf{Novel View Synthesis \textit{in the Wild}.} 
While the above evaluates the quality of the 3D reconstructions, it is not representative of in-the-wild settings where the initial cameras are unknown/approximate and should be optimized during training.
Because even the test camera is approximate, each method is similarly allowed to refine the test camera to better match the test image while keeping the model fixed.
Intuitively, this measures the ability of a model to synthesize a target view under \textit{some} camera. See implementation details in \secref{sec:trained_nvs}.

\textbf{Baselines.} 
We evaluate our approach against Neural Radiance Fields (NeRF)~\citep{mildenhall2020nerf}, which learns a radiance field conditioned on viewpoint and position and renders images using raymarching. We find that the vanilla NeRF struggles in our in-the-wild low-data regime. As such, we make a number of changes to make the NeRF baseline (denoted NeRF$^*$) as competitive as possible, including a mask loss and a canonical volume. Please see the appendix for full details.
We also evaluate a simplified NeRF with a meta-learned initialization for cars from multi-view images~\citep{tancik2020meta}, denoted as MetaNeRF. MetaNeRF meta-learns an initialization such that with just a few gradient steps, it can learn a NeRF model. This allows the model to learn a data-driven prior over the shape of cars. Note that MetaNeRF is trained on ShapeNet~\citep{chang2015shapenet} and thus has seen more data than the other test-time-optimization approaches. We find that the default number of gradient steps was insufficient for MetaNeRF to converge on images from MVMC, so we also evaluate MetaNeRF-ft, which is finetuned until convergence. Finally, we evaluate IDR~\citep{yariv2020multiview}, which represents geometry by extracting a surface from a signed distance field. IDR learns a neural renderer conditioned on the camera direction, position, and normal of the surface.

\textbf{Metrics.}
We evaluate all approaches using the traditional image similarity metrics Mean-Square Error (MSE), Peak Signal-to-Noise Ratio (PSNR), and Structural Similarity Index Measure (SSIM). 
We also compute the Learned Perceptual Image Patch Similarity (LPIPS)~\citep{zhang2018perceptual} which correlates more strongly with human perceptual distance. Finally, we compute the Fr\'echet Inception Distance \citep{heusel2017gans} between the novel view renderings and original images as a measure of visual realism. In \tabref{tab:full_results_fixed} and \tabref{tab:full_results_trained}, we find that NeRS significantly outperforms the baselines in all metrics across both the fixed camera and in-the-wild novel-view synthesis evaluations. See \figref{fig:comparison} and \figref{fig:comparison_trained} for a visual comparison of the methods.

{\bf Qualitative Results.} In \figref{fig:qualitative}, we show qualitative results on our Multi-view Marketplace Cars dataset. Each car instance has between 8 and 16 views. We visualize the outputs of our reconstruction from 3 novel views. We show the rendering for both the full radiance model and the mean texture. Both of these renderings are used to compute the perceptual loss (See \secref{sec:illumination}). We find that NeRS recovers detailed texture information and plausible illumination parameters. To demonstrate the scalability of our approach, we also evaluate on various household objects in \figref{fig:qualitative_misc}. We find that a coarse, cuboid mesh is sufficient as an initialization to recover detailed shape, texture, and lighting conditions. Please refer to the project webpage for 360 degree visualizations.

\section{Discussion}
We present NeRS, an approach for learning neural surface models that capture geometry and surface reflectance. In contrast to volumetric neural rendering, NeRS enforces watertight and closed manifolds. This allows NeRS to model surface-based appearance affects, including view-dependant specularities and normal-dependant diffuse appearance. We demonstrate that such regularized reconstructions allow for learning from sparse in-the-wild multi-view data, enabling reconstruction of objects with diverse material properties across a variety of indoor/outdoor illumination conditions.
Further, the recovery of accurate camera poses in the wild (where classic structure-from-motion fails) remains unsolved and serves as a significant bottleneck for all approaches, including ours. We tackle this problem by using realistic but approximate off-the-shelf camera poses and by introducing a new evaluation protocol that accounts for this.
We hope NeRS inspires future work that evaluates \textit{in the wild} and enables the construction of high-quality libraries of real-world geometry, materials, and environments through better neural approximations of shape, reflectance, and illuminants.

\textbf{Limitations.} 
Though NeRS makes use of factorized models of illumination and material reflectance, there exists some fundamental ambiguities that are difficult from which to recover. For example, it is difficult to distinguish between an image of a gray car under bright illumination and an image of a white car under dark illumination. We visualize such limitations in the supplement. In addition, because the neural shape representation of NeRS is diffeomorphic to a sphere, it cannot model objects of non-genus-zero topologies.

\textbf{Broader Impacts.} NeRS reconstructions could be used to reveal identifiable or proprietary information (e.g., license plates). We made our best effort to blur out all license plates and other personally-identifiable information in the paper and data.

\textbf{Acknowledgements.} We would like to thank Angjoo Kanazawa and Leonid Keselman for useful discussion and Sudeep Dasari and Helen Jiang for helpful feedback on drafts of the manuscript. This work was supported in part by the NSF GFRP (Grant No. DGE1745016), Singapore DSTA, and CMU Argo AI Center for Autonomous Vehicle Research.

\bibliographystyle{plainnat}
\bibliography{bibliography}

\section{Appendix}
\label{sec:appendix}

\begin{figure}
    \centering
    \includegraphics[width=0.8\textwidth]{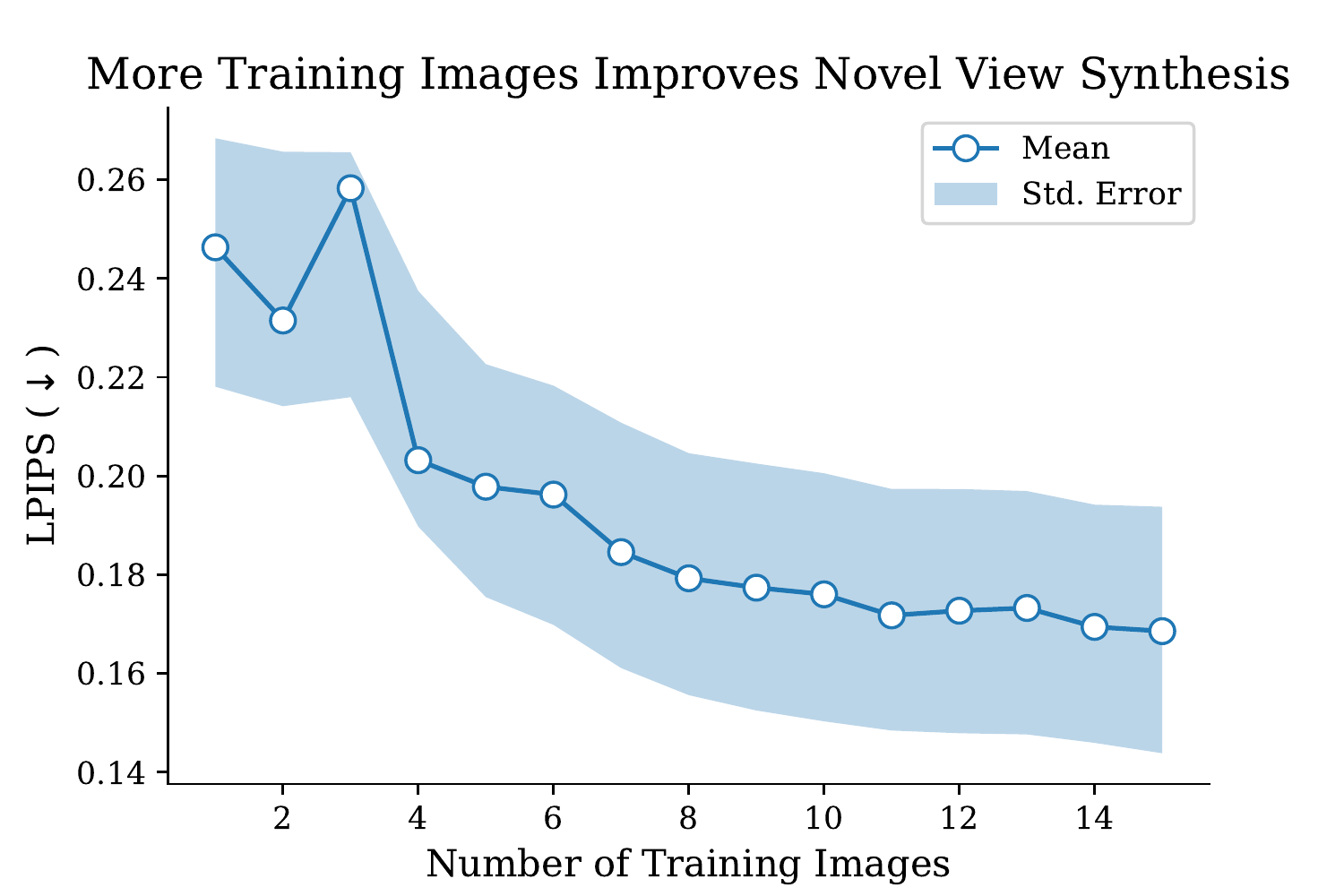}
    \caption{\textbf{Relationship between number of training images and reconstruction quality.} We quantify the number of images needed for a meaningful reconstruction using NeRS on a specific instance from MVMC with 16 total images. Given one of the images as a target image, we randomly select one of the remaining images as the initial training image. Then, we iteratively increase the number of training images, adding the image corresponding to the pseudo-ground truth pose furthest from the current set. This setup is intended to emulate how a user would actually take pictures of objects in the wild (\ie taking wide baseline multi-view images from far away viewpoints).
    Note that once the sets of training images are selected, we use the \textit{in-the-wild} novel view synthesis training and evaluation protocol. In this plot, we visualize the mean and standard error over 16 runs. We find that increasing the number of images improves the quality in terms of perceptual similarity, with performance beginning to plateau after 8 images.}
    \label{fig:num_images}
\end{figure}

\begin{figure}[t]
    \centering
    \includegraphics[width=\textwidth]{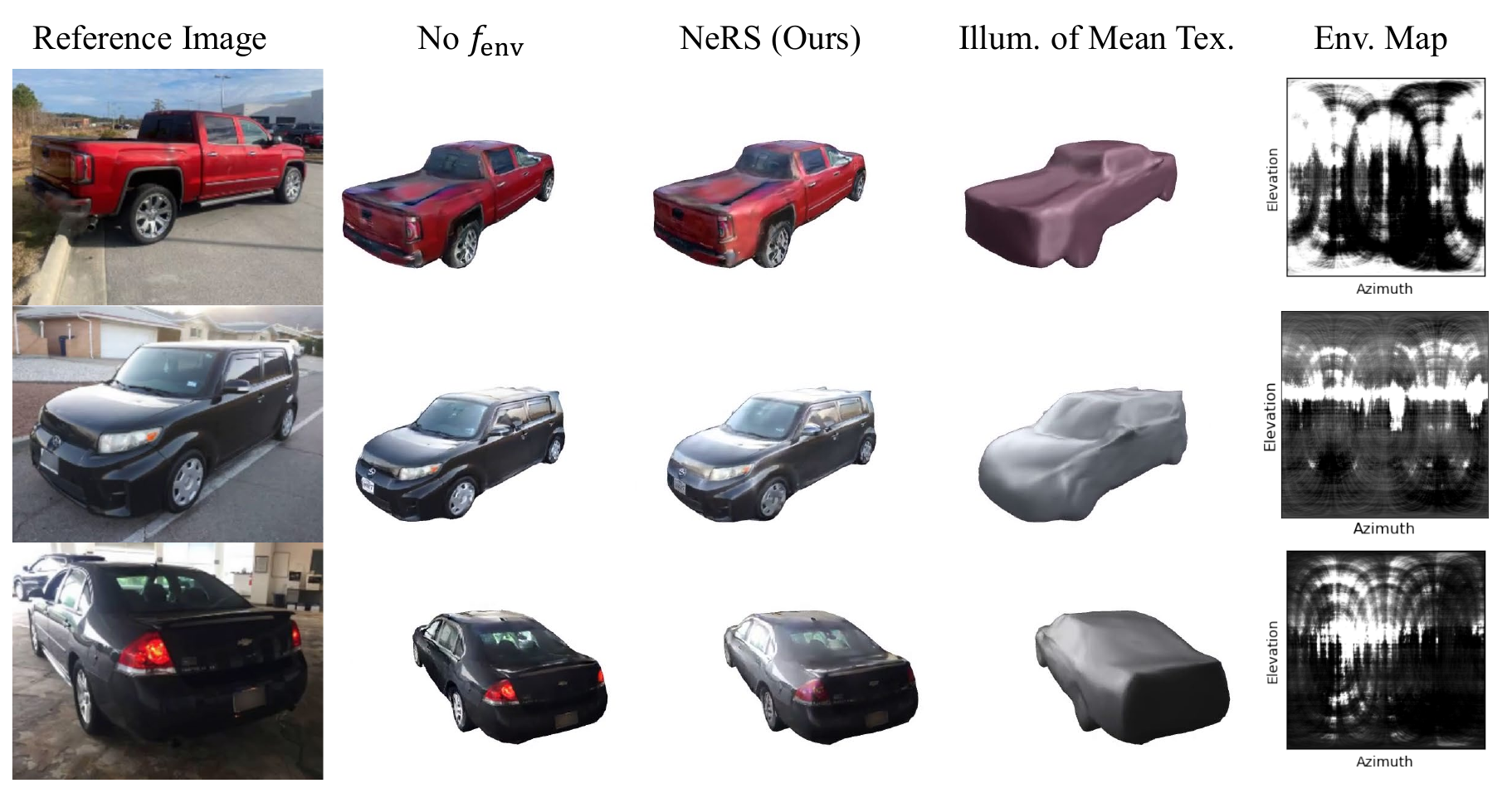}
    \caption{\textbf{Comparison with NeRS trained without view dependence.} Here we compare the full NeRS (column 3) with a NeRS trained without view dependence by only rendering using $\f{tex}$ without $\f{env}$ (column 2). We find that NeRS trained without view-dependence cannot capture lighting effects when they are inconsistent across images. The difference between NeRS trained with and without view-dependence is best viewed in video form. We also visualize the environment map and the illumination of the car with the mean texture. The environment maps show that the light is coming primarily from one side for the first car, uniformly from all directions for the second car, and strongly front left for the third car. \href{https://jasonyzhang.com/ners/paper_figures\#fig10}{[Video]}}
    \label{fig:illumination_ablation}
\end{figure}

\begin{figure}[t!]
    \centering
    \includegraphics[width=\textwidth]{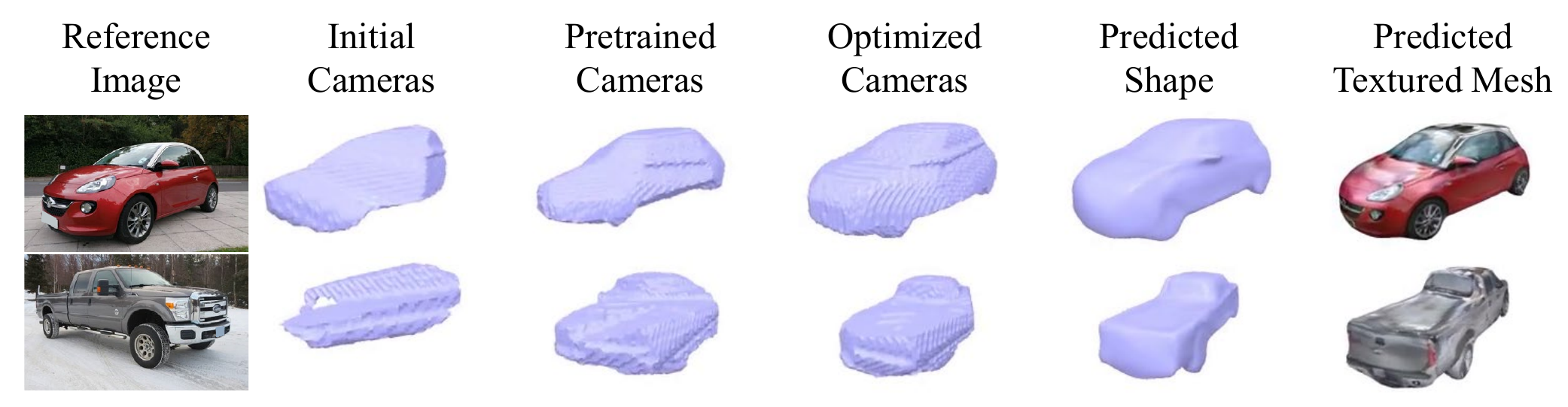}
    \caption{\textbf{Shape from Silhouettes using Volume Carving.} We compare shapes carved from the silhouettes of the training views with the shape model learned by our approach. We construct a voxel grid of size 128$^3$ and keep only the voxels that are visible when projected to the masks using the off-the-shelf cameras (``Initial Cameras"), pre-trained cameras from Stage 1 (``Pretrained Cameras"), and the final cameras after Stage 4 (``Optimized Cameras"). We compare this with the shape model output by $\f{shape}$. We show the nearest neighbor training view and the final NeRS rendering for reference. While volume carving can appear reasonable given sufficiently accurate cameras, we find that the shape model learned by NeRS is qualitatively a better reconstruction. In particular, the model learns to correctly output the right side of the pickup truck and reconstructs the sideview mirrors from the texture cues, suggesting that a joint optimization of the shape and appearance is useful. Also, we note that the more ``accurate" optimized cameras are themselves outputs of NeRS. \href{https://jasonyzhang.com/ners/paper_figures\#fig11}{[Video]}}
    \label{fig:carving}
\end{figure}

\begin{figure}[h]
    \centering
    \includegraphics[width=0.6\textwidth]{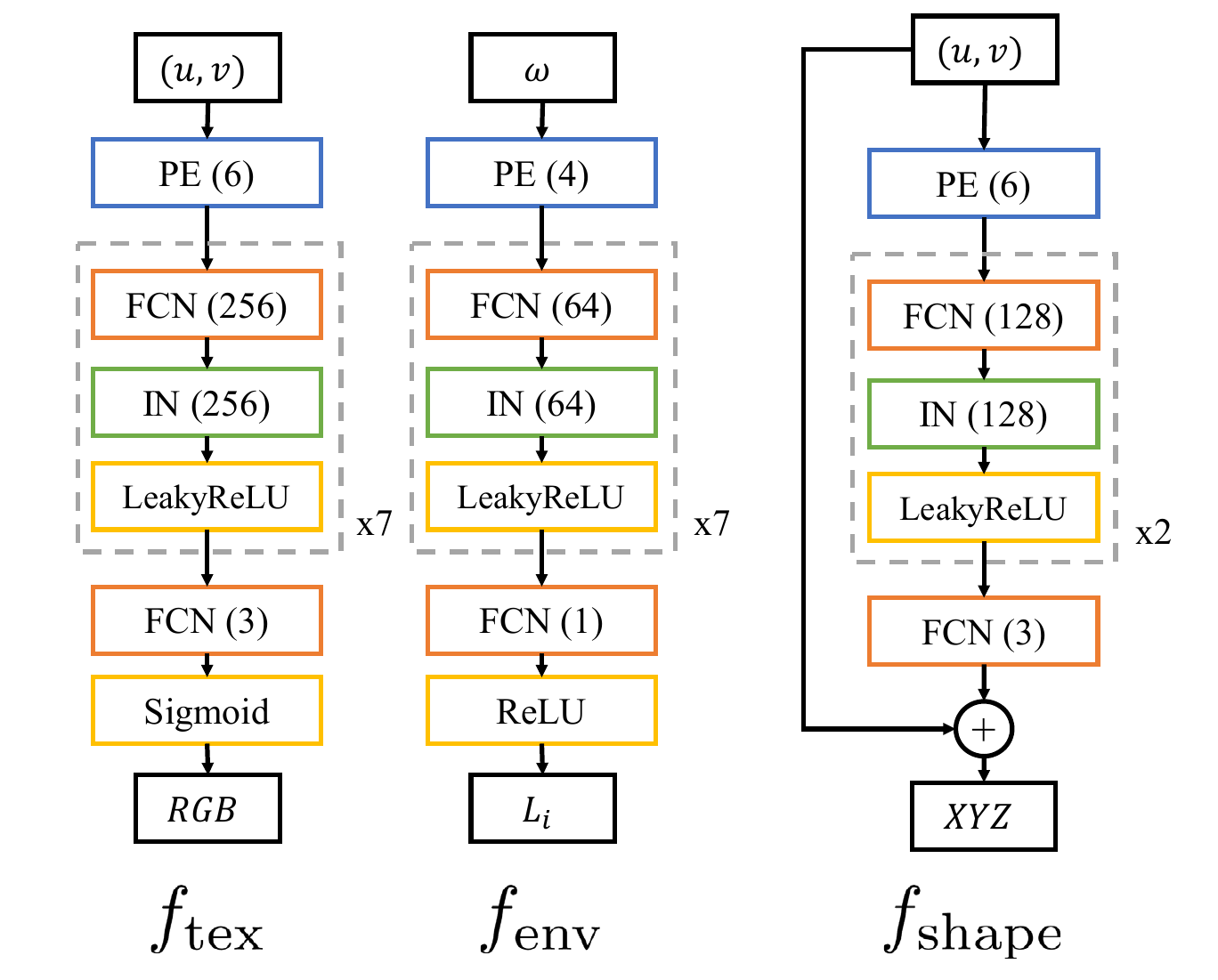}
    \caption{\textbf{Network Architectures.} Here, we show the architecture diagrams for $\f{tex}$, $\f{env}$, and $\f{shape}$. We parameterize $(u,v)$ coordinates as 3-D coordinates on the unit sphere. Following \cite{mildenhall2020nerf,tancik2020fourfeat}, we use positional encoding to facilitate learning high frequency functions, with 6 sine and cosine bases (thus mapping 3-D to 36-D). We use blocks of fully connected layers, instance norm \citep{ulyanov2016instance}, and Leaky ReLU. To ensure texture is between 0 and 1 and illumination is non-negative, we use a final sigmoid and ReLU activation for $\f{tex}$ and $\f{env}$ respectively. We pre-train $\f{shape}$ to output the category-specific mesh template. Given a category-specific mesh, we use an off-the-shelf approach to recover the mesh-to-sphere mapping.\protect\footnotemark}
    \label{fig:architecture}
\end{figure}

We describe an additional ablation on view-dependence in \secref{sec:ablation}, some limitations of our illumination model in \secref{sec:limitations}, implementation details for NeRS and the baselines in \secref{sec:hyperparameters} and \secref{sec:baselines} respectively, and details for in-the-wild novel view synthesis in \secref{sec:trained_nvs}. We also evaluate the effect of increasing the number of training images on novel view synthesis in \figref{fig:num_images}, ablate removing any view-dependence prediction altogether in \figref{fig:illumination_ablation}, compare the learned shape models with volume carving in \figref{fig:carving}, and describe the architectures of NeRS networks in \figref{fig:architecture}.

\subsection{View-Dependence Ablation}
\label{sec:ablation}

To evaluate the importance of learning a BRDF for illumination estimation, we train a NeRS that directly conditions the radiance prediction on the position and view direction, similar to NeRF. More specifically, we concatenate the $uv$ with view direction $\omega$ as the input to $\f{tex}$ which still predicts an RGB color value, and do not use $\f{env}$. We include the quantitative evaluation on using our Multi-view Marketplace Cars (MVMC) dataset for fixed camera novel-view synthesis in \tabref{tab:full_results_fixed_app} and \textit{in-the-wild} novel-view synthesis in \tabref{tab:full_results_trained_app}.

Video qualitative results are available on the figures page\footnote{\url{https://jasonyzhang.com/ners/paper\_figures}} of the project webpage.
We observe that the NeRS with NeRF-style view-direction conditioning has qualitatively similar visual artifacts to the NeRF baseline, particularly large changes in appearance despite small changes in viewing direction. This suggests that learning a BRDF is an effective way to regularize and improve generalization of view-dependent effects.

\begin{table}[t]
    \centering
    \small{
    \begin{tabular}{l c c c c c}
        \toprule
        Method & MSE $\downarrow$ & PSNR $\uparrow$ & SSIM $\uparrow$ & LPIPS $\downarrow$ & FID $\downarrow$ \\
        \cmidrule{1-6}
        NeRF$^*$~\citep{mildenhall2020nerf} & 0.0393 & 16.0 &  0.698 &  0.287 & 231.7\\
        MetaNeRF~\citep{tancik2020meta}& 0.0755 & 11.4 &  0.345 &  0.666 & 394.5\\
        MetaNeRF-ft~\citep{tancik2020meta} & 0.0791 & 11.3  & 0.500 & 0.542 & 326.8 \\
        IDR~\citep{yariv2020multiview} & 0.0698  & 13.8  & 0.658 &  0.328 & 190.1\\
        NeRS (Ours)  & \textbf{0.0254} & \textbf{16.5} &  \textbf{0.720} & \textbf{0.172} & \textbf{60.9}\\
        NeRS + NeRF-style View-dep & 0.0315 & 15.6 &  0.68 & 0.271 & 285.3 \\
        \bottomrule\\
    \end{tabular}
    }

    \caption{\textbf{Quantitative evaluation of novel-view synthesis on MVMC using \textit{fixed pseudo-ground truth cameras}.} 
    Here, we evaluate novel view synthesis with fixed pseudo-ground truth cameras, constructed by manually correcting off-the-shelf cameras that are jointly optimized by our method. In addition to the baselines from the main paper, we compare against an ablation of our approach that directly conditions the radiance on the $uv$ and viewing direction (``NeRS + NeRF View-dep") in a manner similar to NeRF.}
    \vspace{-5mm}
    \label{tab:full_results_fixed_app}
\end{table}

\begin{table}[t!]
    \centering
    \small{
    \begin{tabular}{p{2cm} l c c c c c}
        \toprule
        Target Camera & Method & MSE $\downarrow$ & PSNR $\uparrow$ & SSIM $\uparrow$ & LPIPS $\downarrow$ & FID $\downarrow$ \\
        \cmidrule{1-7}
        \multirow{4}{\linewidth}{\centering Approx. Camera}
 & NeRF$^*$~\citep{mildenhall2020nerf} & 0.0657 & 12.7 & 0.604 & 0.386 &  290.7\\
        & IDR~\citep{yariv2020multiview} & 0.0836 & 11.4 &  0.591 & 0.413 &  251.6 \\
        & NeRS (Ours)  & 0.0527 & 13.3 &  0.620 & 0.284  & 100.9 \\
        & NeRS + NeRF-style View-dep. & 0.0573 & 13.0 & 0.624 & 0.356  & 296.4 \\
        \cmidrule(lr){1-7}

        \multirow{4}{\linewidth}{\centering Refined Approx. Camera}
 & NeRF$^*$~\citep{mildenhall2020nerf} & 0.0464 & 14.7 &  0.660 &  0.335 & 277.9\\
        & IDR~\citep{yariv2020multiview} & 0.0454 & 14.4  & 0.685 &  0.297 & 242.3\\
        & NeRS (Ours)  & \textbf{0.0338} & \textbf{15.4} &  \textbf{0.675} & \textbf{0.221} & \textbf{92.5}\\
        & NeRS + NeRF-style View-dep. & 0.0482 & 13.9 & 0.659  & 0.316 &  293.5 \\
        \bottomrule\\
    \end{tabular}
    }
    \vspace{-2mm}
    \caption{\textbf{Quantitative evaluation of \textit{in-the-wild} novel-view synthesis on MVMC.}
    To evaluate \textit{in-the-wild} novel view synthesis, each method is trained with (and can refine) approximate cameras estimated using \cite{Xiao2019PoseFromShape}. To compensate for the approximate test camera, we allow each method to refine the test camera given the target image. In this table, we show the performance before (``Approx. Camera") and after (``Refined Approx. Camera") this refinement. All methods improve from the refinement. In addition, we compare against an ablation of our approach that directly conditions the radiance on the $uv$ and viewing direction (``NeRS + NeRF View-dep") in a manner similar to NeRF.
    }
    \vspace{-5mm}
    \label{tab:full_results_trained_app}
\end{table}

\subsection{Limitations of illumination model}
\label{sec:limitations}

\begin{figure}[h]
    \centering
    \includegraphics[width=\textwidth]{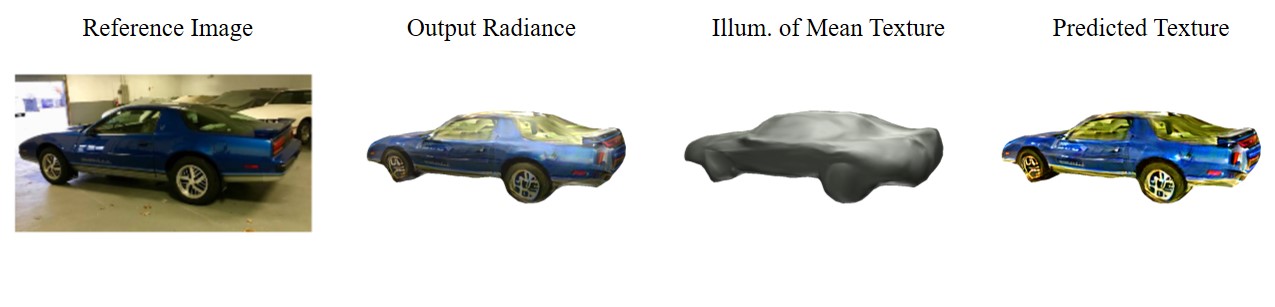}
    \caption{\textbf{Non-white lighting.} Our illumination model assumes environment lighting to be grayscale. In this car, the indoor lighting is yellow, which causes the yellow lighting hue to be baked into the predicted texture (Predicted Texture) rather than the illumination (Illum. of Mean Texture).}
    \label{fig:non_white_illumination}
\end{figure}

\begin{figure}[h]
    \centering
    \includegraphics[width=\textwidth]{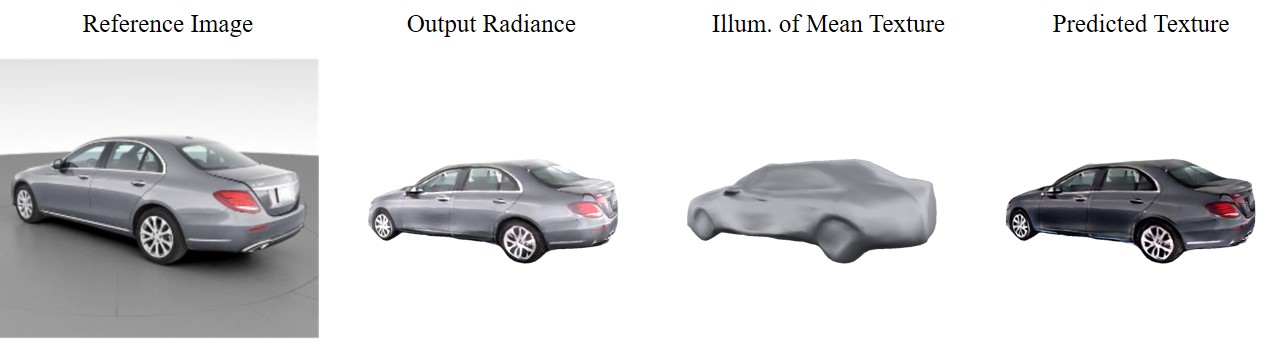}
    \caption{\textbf{Gray texture-illumination ambiguity.} For gray cars, there is a brightness ambiguity between texture and lighting. Although the car shown here is silver, another possible interpretation is that the car is dark grey with very bright illumination and specularity.}
    \label{fig:gray_ambiguity}
\end{figure}

In order to regularize lighting effects, we assume all lighting predicted by $\f{env}$ to be grayscale. For some images with non-white lighting (see \figref{fig:non_white_illumination}), this can cause the lighting color to be baked into the predicted texture in an unrealistic manner. In addition, if objects are gray in color, there is a fundamental ambiguity as to whether the luminance is due to the object texture or lighting \eg dark car with bright illumination or light car with dark illumination (see \figref{fig:gray_ambiguity}).

\subsection{Implementation Details and Hyper-Parameters}
\label{sec:hyperparameters}

To encourage plausible lighting, we found it helpful to also optimize a perceptual loss on the illumination of the mean texture, similar to \cite{wu2021derender}. During optimization, we render an image $\hat{I}$ with full texture using $\f{tex}$ as well as a rendering $\bar{I}$ using the mean output of $\f{tex}$. We compute a perceptual loss on both $\hat{I}$ and $\bar{I}$. We weight these two perceptual losses differently, as shown in \tabref{tab:loss_weights}.
To compute illumination, we sample environment rays uniformly across the unit sphere, and compute the normals corresponding to each vertex, ignoring rays pointing in the opposite direction of the normal.

\subsection{Implementation Details of Baselines}
\label{sec:baselines}

\textbf{NeRF}\footnote{\url{https://github.com/facebookresearch/pytorch3d/tree/main/projects/nerf}}~\citep{mildenhall2020nerf}: We find that a vanilla NeRF struggles when given sparse views. As such, we make the following changes to make the NeRF baseline as competitive as possible: 1. we add a mask loss that forces rays to either pass through or be absorbed entirely by the neural volume, analogous to space carving~\citep{kutulakos2000theory} (see \figref{fig:carving}); 2. a canonical volume that zeros out the density outside of a tight box where the car is likely to be. This helps avoid spurious ``cloudy" artifacts from novel views; 3. a smaller architecture to reduce overfitting; and 4. a single rendering pass rather than dual coarse/fine rendering passes. In the main paper, we denote this modified NeRF as NeRF$^*$. For in-the-wild novel-view synthesis, we refine the training and test cameras directly, similar to \cite{wang2021nerf}.

\textbf{MetaNeRF}\footnote{\url{https://github.com/tancik/learnit}}~\citep{tancik2020meta}: We fine-tuned the pre-trained 25-view ShapeNet model on our dataset. We used the default hyper-parameters for the MetaNeRF baseline. For MetaNeRF-ft, we increased the number of samples to 1024, trained with a learning rate of 0.5 for 20,000 iterations, then lowered the learning rate to 0.05 for 50,000 iterations.

\footnotetext{\url{https://github.com/icemiliang/spherical_harmonic_maps}}
\textbf{IDR}\footnote{\url{https://github.com/lioryariv/idr}}~\citep{yariv2020multiview}: Because each instance in our dataset has fewer images than DTU, we increased the number of epochs by 5 times and adjusted the learning rate scheduler accordingly. IDR supports both fixed cameras (which we use for \textit{fixed} camera novel-view synthesis) and trained cameras (which we use for \textit{in-the-wild} novel-view synthesis).

\begin{table}[t]
    \centering
    \begin{tabular}{c c c c c c c c c}
        \toprule
        \multirow{2}{*}{Stage} & \multirow{2}{2cm}{\centering Optimize} & \multirow{2}{1cm}{\centering Num. Iters}  & \multicolumn{6}{c}{Loss Weights}   \\
            \cmidrule{4-9}
             &  &  & Mask & DT & Chamfer & Perc. & Perc.~mean & Reg.\\
        \midrule
        1 & $\{\Pi_i\}_{i=1}^N$ & 500  & 2 & 20 & 0.02 & 0 & 0 & 0\\
        2 & Above + $\f{shape}$ & 500 & 2 & 20 & 0.02 & 0 & 0 & 0.1 \\
        3 & Above + $\f{tex}$ & 1000 & 2 & 20 & 0.02 & 0.5 & 0 & 0.1\\ 
        4 & Above + $\f{env}, \alpha, k_s$ & 500 & 2 & 20 & 0.02 & 0.5 & 0.15 & 0.1\\ 
        \bottomrule \\
    \end{tabular}
    \caption{\textbf{Multi-stage optimization loss weights and parameters.} We employ a 4 stage training process: first optimizing just the cameras before sequentially optimizing the shape parameters, texture parameters, and illumination parameters. In this table, we list the parameters optimized during each stage as well as the number of training iterations and loss weights. All stages use the Adam optimizer~\citep{kingma2014adam}. ``Perc.~refers" to the perceptual loss on the rendered image with full texture while ``Perc.~mean" refers to the perceptual loss on the rendered image with mean texture.}
    \label{tab:loss_weights}
\end{table}

\subsection{\textit{In-the-wild} Novel-view Synthesis Details}
\label{sec:trained_nvs}

For \textit{in-the-wild} novel-view synthesis without ground truth cameras, we aim to evaluate the capability of each approach to recover a meaningful 3D representation while only using approximate off-the-shelf cameras. 
Given training images with associated approximate cameras, each method is required to output a 3D model.
We then evaluate whether this 3D model can generate a held-out target image under \textit{some} camera pose.

In practice, we recover approximate cameras using \cite{Xiao2019PoseFromShape}. However, as these only comprise of a rotation, we further refine them using the initial template and foreground mask. The initial approximate cameras we use across all methods are in fact equivalent to those obtained by our method after Stage 1 of training, and depend upon: i) prediction from \cite{Xiao2019PoseFromShape}, and ii) input mask and template shape. During training, each method learns a 3D model while jointly optimizing the camera parameters using gradient descent. Similarly, at test time, we refine approximate test camera with respect to the training loss function for each method (\ie \eref{eq:total_loss} for NeRS, MSE for NeRF, and RGB+Mask loss for IDR) using the Adam optimizer for 400 steps. We show all metrics before and after this camera refinement in \tabref{tab:full_results_trained_app}.

\end{document}